\documentclass[10pt,twocolumn,letterpaper]{article}

\usepackage[pagenumbers]{cvpr} 

\usepackage{graphicx}
\usepackage{amsmath}
\usepackage{amssymb}
\usepackage{booktabs}

\usepackage{times}
\usepackage{epsfig}

\makeatletter
\@namedef{ver@everyshi.sty}{}
\makeatother
\usepackage{tikz}
\usepackage{pgfplots}
\pgfplotsset{compat=1.7}
\usetikzlibrary{shapes.geometric,decorations.pathreplacing,arrows.meta,calc,matrix,positioning}
\usepackage[inline]{enumitem}
\usepackage{appendix}

\usepackage{bm}

\usepackage[pagebackref,breaklinks,colorlinks]{hyperref}

\usepackage[capitalize]{cleveref}
\crefname{section}{Sec.}{Secs.}
\Crefname{section}{Section}{Sections}
\Crefname{table}{Table}{Tables}
\crefname{table}{Tab.}{Tabs.}

\def\cvprPaperID{11200} 
\def\confName{CVPR}
\def\confYear{2022}

\begin{document}
\def \ModelName {X-Pool}

\title{\ModelName{}: Cross-Modal Language-Video Attention for Text-Video Retrieval}

\author{Satya Krishna Gorti\textsuperscript{1}\thanks{Authors contributed equally to this work.} \hspace{1cm} Noël Vouitsis\textsuperscript{1,2}\footnotemark[1] \hspace{1cm} Junwei Ma\textsuperscript{1}\footnotemark[1]\\
Keyvan Golestan\textsuperscript{1} \hspace{1cm} Maksims Volkovs\textsuperscript{1} \hspace{1cm} Animesh Garg\textsuperscript{2,3,4} \hspace{1cm} Guangwei Yu\textsuperscript{1}\\\\
\textsuperscript{1}Layer 6 AI \hspace{0.25cm} \textsuperscript{2}University of Toronto \hspace{0.25cm} \textsuperscript{3}Vector Institute \hspace{0.25cm} \textsuperscript{4}NVIDIA\\
}

\maketitle

\begin{abstract}
In text-video retrieval, the objective is to learn a cross-modal similarity function between a text and a video that ranks relevant text-video pairs higher than irrelevant pairs. However, videos inherently express a much wider gamut of information than texts. Instead, texts often capture sub-regions of entire videos and are most semantically similar to certain frames within videos.
Therefore, for a given text, a retrieval model should focus on the text's most semantically similar video sub-regions to make a more relevant comparison. Yet, most existing works aggregate entire videos without directly considering text. Common text-agnostic aggregations schemes include mean-pooling or self-attention over the frames, but these are likely to encode misleading visual information not described in the given text. To address this, we propose a cross-modal attention model called \ModelName{} that reasons between a text and the frames of a video. Our core mechanism is a scaled dot product attention for a text to attend to its most semantically similar frames. We then generate an aggregated video representation conditioned on the text's attention weights over the frames. We evaluate our method on three benchmark datasets of MSR-VTT, MSVD and LSMDC, achieving new state-of-the-art results by up to 12\% in relative improvement in Recall@1. Our findings thereby highlight the importance of joint text-video reasoning to extract important visual cues according to text. Full code and demo can be found at: \href{https://layer6ai-labs.github.io/xpool/}{\textbf{layer6ai-labs.github.io/xpool/}}.

\end{abstract}

\section{Introduction}
\begin{figure}[t]\centering
           \includegraphics[width=\columnwidth]{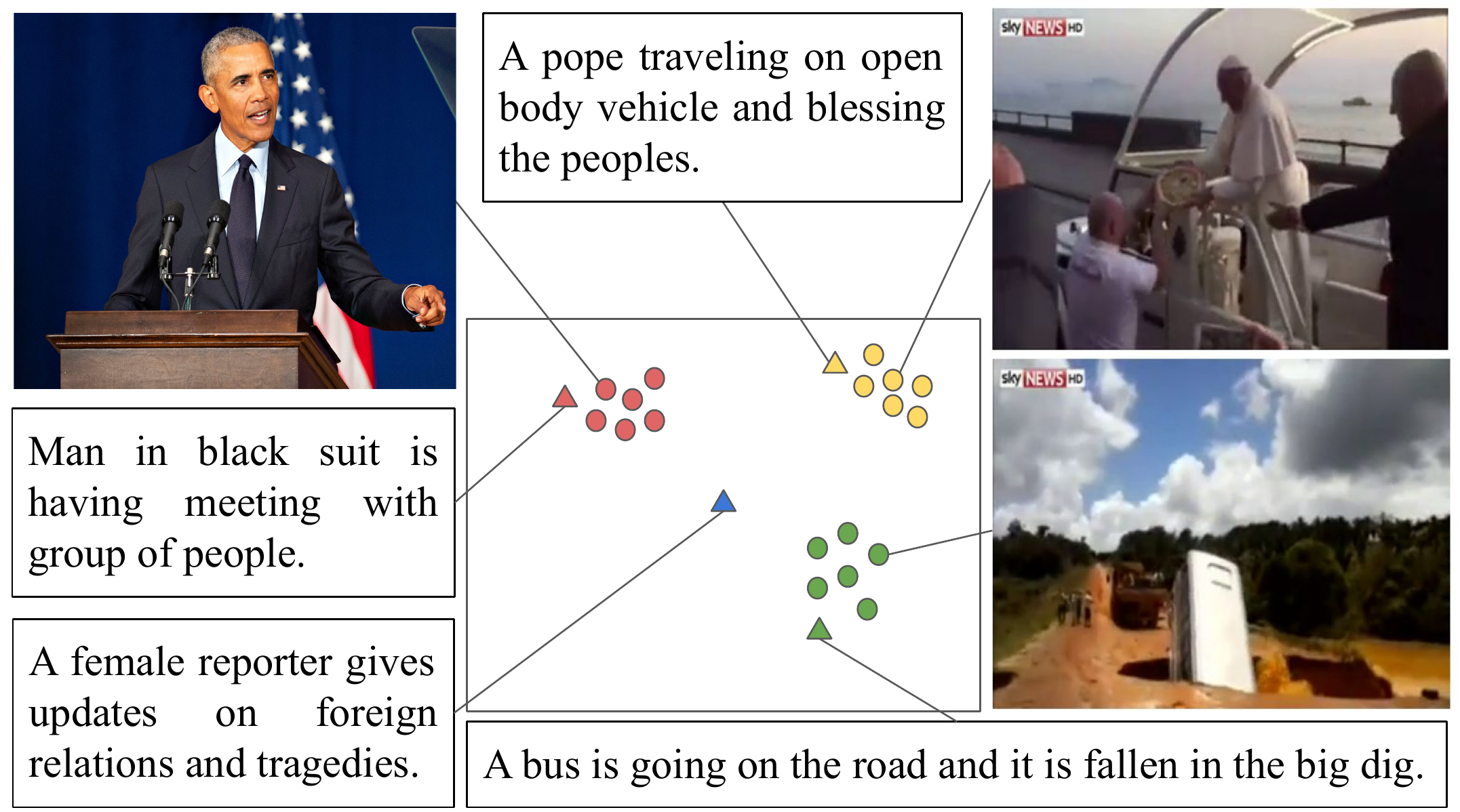}\\
  \caption{Illustration of the joint text and visual representations for a single video and its captions taken verbatim from the MSR-VTT dataset. Since the video is capturing more content than each individual text, aggregating the entire video regardless of the input text can be misleading.}
\label{fig:examples}
\vspace{-0.4cm}
\end{figure}
The advent of video content platforms like TikTok, YouTube and Netflix have enabled the mass outreach of videos around the world.
The ability to retrieve videos that are most semantically similar to a provided text-based search query allows us to quickly find relevant information and to make sense of massive amounts of video data.

The task of text-video retrieval is an approach to solve this problem wherein the objective is for a model to learn a similarity function between texts and videos. To compute the similarity between both modalities, a common technique is to first embed a text and a video into a joint latent space and then apply a distance metric such as the cosine similarity between the text and video embeddings
\cite{liu2019use, gabeur2020multi, bain2021frozen}.

However, there is an important discrepancy between both modalities that makes such a direct comparison challenging.
Videos inherently express a much wider gamut of information than texts, so a text generally cannot fully capture the entire contents of a video. Instead, texts are most semantically similar to sub-regions of videos, represented as a subset of frames. Depending on the given text, the frames that are the most semantically similar would differ, so multiple equally valid texts can match a particular video. For example, in Figure \ref{fig:examples}, we show frames of a sample video from the MSR-VTT dataset \cite{xu2016msr}. The frames depict various scenes from international news and express different visual content. Moreover, we show multiple captions associated with this video, and observe that each caption best matches a different video frame but can seem irrelevant to others. In this example, we would expect the same video to be retrieved for any of these queries, even though the relevant content is limited to sub-regions of the video.

Based on this observation, we want a retrieval model to focus on the video sub-regions that are most relevant to the given text during retrieval. A model should therefore directly reason between texts and the frames of videos to extract the most relevant information as described in each text. However, most existing works do not apply direct cross-modal reasoning, and instead utilize the entire contents of a video such as through mean-pooling or self-attention \cite{miech2019howto100m, gabeur2020multi, bain2021frozen, luo2021clip4clip}. By encoding a video independently from a given text, a model is likely to encode superfluous or even distracting visual information that is not described in the text, which can reduce retrieval performance.

To address this gap, we design a cross-modal attention model that we call \textbf{\ModelName{}} to allow for joint reasoning between a text and a video's frames. Unlike previous works that pool the entire frames of a video, our model provides flexibility for a text to attend to its most semantically similar frames and then generates an aggregated video representation conditioned on those frames.

Our main contributions can be summarized as follows: \begin{enumerate*}[label=(\roman*)]
\item We show empirically through a proof of concept that text-conditioned video pooling allows a model to reason about the most relevant video frames to a given text, which outperforms baselines that use text-agnostic video pooling;
\item We propose a cross-modal attention model that extends our proof of concept with parametric capacity for a text to attend to its most semantically similar video frames for aggregation which we call \ModelName{}. \ModelName{} obtains state-of-the-art results across the popular benchmark datasets of MSR-VTT \cite{xu2016msr}, MSVD \cite{chen2011collecting} and LSMDC \cite{rohrbach2017movie};

\item We demonstrate the robustness of \ModelName{} to videos with increasing amounts of content diversity, such as videos with many scene transitions. We show how text-agnostic pooling methods are much more sensitive to such videos  compared to our text-conditioned \ModelName{} model.

\end{enumerate*}

\section{Related Work}

\textbf{Joint Language-Image Understanding.} 

Joint language-image models are a form of multimodal learning \cite{baltruvsaitis2018multimodal} that aim to understand and relate the text and image modalities. Methods in text-image understanding such as \cite{li2019visualbert, lu2019vilbert, tan2019lxmert, chen2020uniter, li2020oscar, radford2021learning, jia2021scaling, li2021align} are pre-trained to jointly reason about language and image semantics which make them suitable for downstream cross-modal tasks like visual question answering (VQA) \cite{antol2015vqa}, image captioning \cite{xu2015show} and text-image retrieval \cite{karpathy2015deep}. Most recently, methods such as CLIP \cite{radford2021learning}, ALIGN \cite{jia2021scaling}, DeCLIP \cite{li2021supervision} and ALBEF \cite{li2021align} employ unimodal encoders to learn a joint latent space that matches relevant text-image pairs via a contrastive loss. Our goal is to bootstrap from a pre-trained joint text-image model and extend it towards a joint text-video model for the task of text-video retrieval.

\textbf{Text-Video Retrieval.} The prototypical approach to text-video retrieval has been through a pre-trained language expert and often a combination of video experts pre-trained for various tasks and modalities, after which the language and vision streams are consolidated through late fusion. MoEE \cite{miech2018learning}, CE \cite{liu2019use}, MMT \cite{gabeur2020multi} MDMMT \cite{dzabraev2021mdmmt}, and TeachText \cite{croitoru2021teachtext} are all such works. The motivation for using pre-trained experts stems from the small-scale nature of the datasets used in text-video retrieval.

Some works have also benefited from pre-training their own models on either large-scale text-video datasets \cite{miech2019howto100m, zhu2020actbert, bain2021frozen} or through text-image pre-training \cite{miech2018learning, lei2021less}. Among them, ActBERT \cite{zhu2020actbert} and ClipBERT \cite{lei2021less} are both single
stream models that jointly embed text-video pairs through BERT-like architectures for early cross-modal fusion. However, these works do not allow for direct reasoning about the most semantically similar video sub-regions to a given text.

Recently, the works of CLIP4Clip \cite{luo2021clip4clip} and Straight-CLIP \cite{portillo2021straightforward} use the joint language-vision model of CLIP \cite{radford2021learning} pre-trained on a large-scale text-image dataset as a backbone. Even the trivial use of CLIP in a zero-shot manner outperforms most of the above recent works \cite{portillo2021straightforward}, highlighting how the rich joint text-image understanding of CLIP can be expanded towards videos. CLIP4Clip \cite{luo2021clip4clip} proposes several video aggregation schemes including mean-pooling, self-attention and a multimodal transformer, yet none allow for direct matching of a text with its most relevant video sub-regions which motivates our cross-modal attention model. Cross-modal attention has been explored in previous related work such as \cite{yao2015describing, li2019visualbert, lu2019vilbert, tan2019lxmert, chen2020uniter, li2020oscar, zhu2020actbert, lei2021less, miech2021thinking, zhang2021temporal, tan2021look, li2021align}. We design a cross-modal attention mechanism for the task of text-video retrieval that shows significant improvement over previous methods.

\section{Problem Statement}
In text-video retrieval, the objective is for a model to learn a scalar similarity function $s(t, v)$ between a text $t$ and a video $v$. We want to assign higher similarity to relevant text-video pairs and assign lower similarity to irrelevant pairs.
We define two retrieval tasks, text-to-video retrieval denoted as $t2v$ and video-to-text retrieval denoted as $v2t$. In $t2v$, we are given a query text $t$ and a video index set $\mathcal{V}$. The goal is to rank all videos $v \in \mathcal{V}$ according to their similarities with the query text.  Analogously, in $v2t$, we are given a query video $v$ and a text index set $\mathcal{T}$. The goal is to rank all texts $t \in \mathcal{T}$ according to their similarities with the query video. In both of these tasks, we are under the assumption that only the index set is known ahead of time.

The inputs to our problem are a video $v$ and a text $t$. We define a video $v \in \mathbb{R}^{F \times 3 \times H \times W}$ as a sequence of $F$ sampled image frames in time. That is, $v = [v^{1}, v^{2}, \cdots, v^F]^T$ where $v^{f}$ is the $f^{\text{th}}$ image frame of resolution $H \times W$. We define a text $t$ as a sequence of tokenized words.

\section{Methodology}

In this section, we incrementally introduce the insights and methodologies that motivate our final model \ModelName{}. We first describe in Section \ref{sec:clip} how the use of a pre-trained joint text-image model is an essential component of our model to match texts and images which we extend to match texts and videos. We then explain the drawbacks of aggregating a video into a text-agnostic embedding in Section \ref{sec:no-text-pool}, and present an alternative framework that aggregates frames conditioned on a given text in Section \ref{sec:text-cond-pool}. We then introduce our \ModelName{} model in Section \ref{sec:our-model}, a cross-modal attention model that enables joint reasoning between a text and the frames of a video. Our model learns to aggregate videos using the most semantically similar frames to a given text.

\begin{figure*}[t]\centering
           \includegraphics[width=1\textwidth]{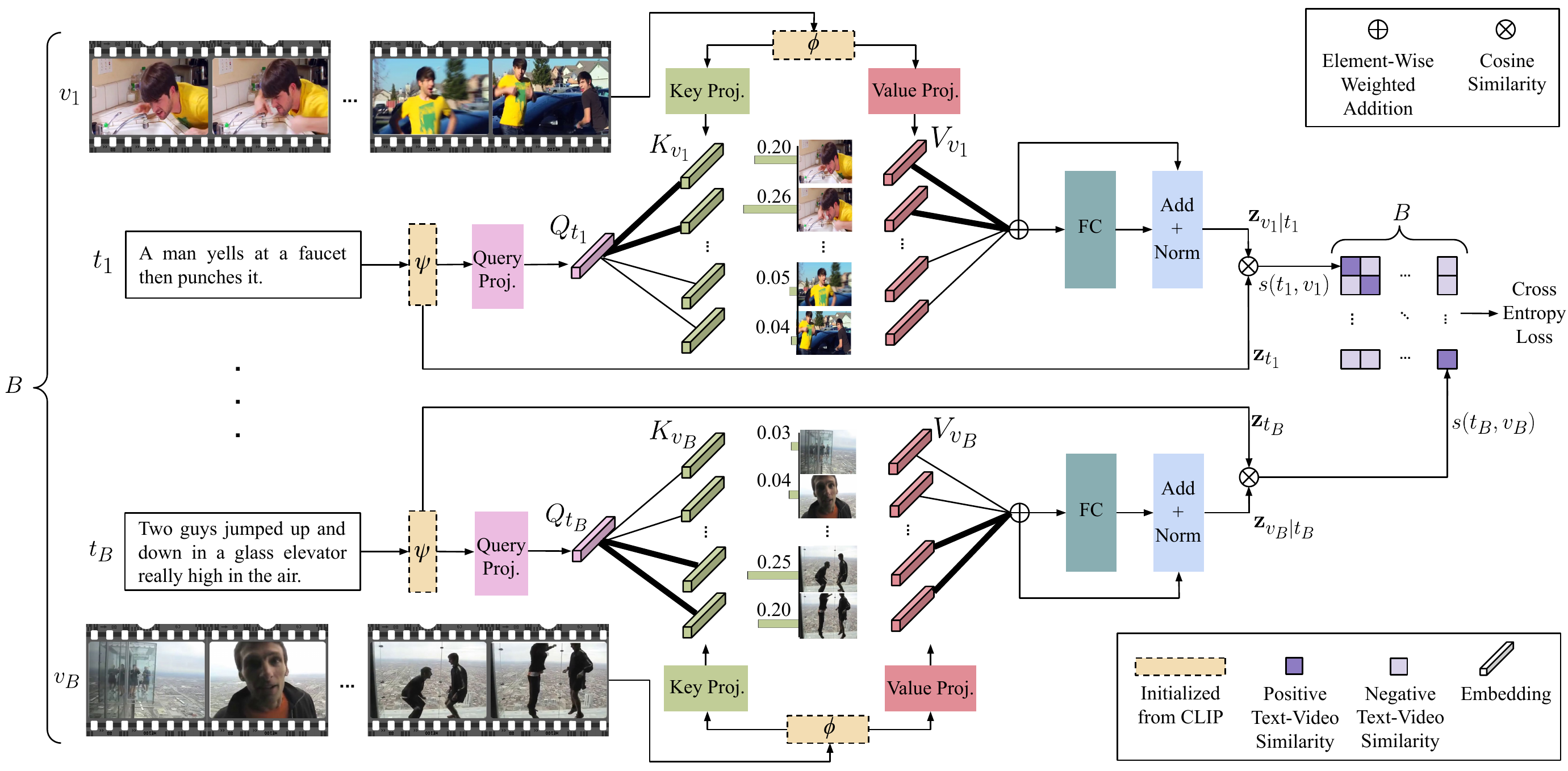}\\
  \caption{Diagram of \ModelName{}. For the given text $t_1$, we embed it with the text encoder $\psi$ and then apply a query projection to obtain $Q_{t_1}$. We similarly embed the frames of the given video $v_1$ with the image encoder $\phi$ and then apply a key projection to obtain $K_{v_1}$. We compute the dot product attention between them as illustrated by the horizontal bar plot in the middle of the figure. Our attention mechanism allows \ModelName{} to focus on the most relevant frames given an input text. We aggregate a separate set of value-projected frame embeddings that we weight by the previously computed dot product attention scores to obtain an aggregated video embedding that we then pass through a fully connected layer (FC) with a residual connection to obtain $\textbf{z}_{v_1|t_1}$. We compute the similarity score $s(t_1,v_1)$ as the cosine similarity between $\textbf{z}_{v_1|t_1}$ and $\textbf{z}_{t_1}=\psi(t_1)$. Finally, we compute a cross entropy loss after obtaining $s(t_i, v_j)$ as just described for each pair $(t_i, v_j)$ within a batch of size $B$.
} 
  \label{fig:arch}
  \vskip -0.2cm
\end{figure*}

\subsection{Expanding Joint Text-Image Models}
\label{sec:clip}
\textbf{Bootstrapping From Joint Text-Image Models.} Jointly pre-trained text-image models have demonstrated the ability to match semantically similar texts and images \cite{lu2019vilbert, chen2020uniter, jia2021scaling, radford2021learning, li2021supervision, li2021align}. We can leverage the existing text-image reasoning of such models to bootstrap a joint text-video model. This allows us to learn language-video interactions with substantially less video data and offers a more compute efficient solution during training, while benefiting from the rich cross-modal understanding of pre-trained joint text-image models. In general, the idea of bootstrapping video models from image models stems from the importance of first understanding images in order to understand videos, as shown in \cite{carreira2017quo}. 

\textbf{CLIP as a Backbone.} We bootstrap from CLIP \cite{radford2021learning} due to its strong downstream performance, its simplicity, and to more objectively compare with recent works that also leverage CLIP as a backbone \cite{portillo2021straightforward, luo2021clip4clip}, although other pre-trained text-image models may be suitable backbone candidates. 
To bootstrap from CLIP for text-video retrieval, we first embed a text and individual video frames into its joint latent space and then pool the frame embeddings to obtain a video embedding \cite{portillo2021straightforward}. Since the existing information extracted from a pre-trained CLIP model contains rich text-image semantics, we use CLIP as a backbone to learn a new joint latent space to match texts and videos instead of just images.

More precisely, given a text $t$ a video frame $v^f$ as input, CLIP outputs a text embedding $\mathbf{c}_t \in \mathbb{R}^D$ and a frame embedding $\mathbf{c}_v^f \in \mathbb{R}^D$ in a joint latent space: 
\begin{align}
    \mathbf{c}_t &= \psi(t)\\
    \mathbf{c}_v^f &= \phi(v^f)\label{eq:clip-img}
\end{align}
where $\psi$ is CLIP's text encoder and $\phi$ is CLIP's image encoder. By computing equation \eqref{eq:clip-img} for each frame in a video $v$, we obtain a sequence of frame embeddings $C_v = [\mathbf{c}_v^1, \mathbf{c}_v^2, \cdots, \mathbf{c}_v^F]^T \in \mathbb{R}^{F \times D}$.

\textbf{Computing Text and Video Embeddings.} As mentioned, we want to embed our given text and video into a joint space to compute similarity. That is, we want to compute a text embedding $\mathbf{z}_t \in \mathbb{R}^D$ and a video embedding $\mathbf{z}_v \in \mathbb{R}^D$. The text embedding is directly taken as the output from CLIP. On the other hand, we compute the video embedding by aggregating the frame embeddings in $C_v$ using a temporal aggregation function $\rho$:
\begin{align}
    \mathbf{z}_{t} &= \mathbf{c}_{t}\\
    \mathbf{z}_{v} &= \rho(C_v)
\end{align}

\subsection{Gap: Text-Agnostic Pooling}
\label{sec:no-text-pool}
In most existing works, the aggregation function $\rho$ does not directly consider the input text and is purely a function of the frames of the videos such as through mean-pooling, self-attention or an LSTM \cite{miech2019howto100m, gabeur2020multi, miech2020end, alayrac2020self, bain2021frozen, portillo2021straightforward, luo2021clip4clip}.

While defining the temporal aggregation function as agnostic to text forms a simple baseline, there are important drawbacks with this approach. Videos are inherently much more expressive than texts, so the information captured in text generally cannot fully capture that of an entire video. Instead, texts are most semantically similar to certain sub-regions of videos which we define as subsets of frames, as shown in Figure \ref{fig:examples}. As such, common text-agnostic aggregation schemes that pool entire videos like mean-pooling and self-attention might encode spurious information that is not described in the input text.

We note that this effect is exacerbated when we consider videos that exhibit significant diversity in their visual content \cite{liu2009recognizing} which we refer to as content diversity. To elaborate, it is natural to find videos with scene transitions such as when the actor moves from an indoor setting to an outdoor setting, abrupt scene cuts like in movies, occlusions of key subjects or noise in the form of distractors for example. Since this is an intrinsic property of many videos ``in the wild'' \cite{soomro2012ucf101, li2019quality}, we want a retrieval model to be robust to such content diversity by focusing its attention to the most relevant video sub-regions described in a given text. Intuitively, any text-agnostic pooling method will fail under this setting since it aggregates information from all scenes of the video, disregarding the input text for retrieval, as we empirically show in Section \ref{sec:results}.

\subsection{Key Insight: Text-Conditioned Pooling}
\label{sec:text-cond-pool}
We note that it is therefore important to match texts not with the entire contents of a video, but with those video frames that are most semantically similar to a given text. Depending on the given text, the frames that are most semantically similar would differ, so there could be multiple equally valid texts that match a particular video. As such, our temporal aggregation function should directly reason between a given text and the frames of a video.

To that end, we formulate a new temporal aggregation function $\pi$ that allows us to aggregate the video frames that are most semantically similar to a given text $t$. By conditioning $\pi$ on $t$, we can extract from a video $v$ the most relevant information as described in $t$ while suppressing noisy and misleading visual cues. We denote the resulting aggregated video embedding as $\mathbf{z}_{v \mid t}$ and define our similarity function $s(t, v)$ as:

\begin{equation}
    \mathbf{z}_{v \mid t} = \pi(C_v \mid t)
\end{equation}
\begin{equation}
    \label{eq:cosine-sim-new}
    s(t, v) = \frac{\mathbf{z}_t \cdot \mathbf{z}_{v \mid t}}{\|\mathbf{z}_t\| \|\mathbf{z}_{v \mid t}\|}
\end{equation}

To demonstrate the efficacy of our idea, we first propose a top-$k$ aggregation function $\pi_{\text{top-}k}(C_v \mid t)$ as:
\begin{equation}
   \pi_{\text{top-}k}(C_v \mid t) = \frac{1}{k}\sum_{f \in \mathcal{K}} \mathbf{c}_v^{f}
   \label{eq:topk}
\end{equation}
where the set $\mathcal{K}$ is defined as:
\begin{align}
    \mathcal{K} &= \underset{\substack{\mathcal{K}\subseteq\{1,...,F\}\\|\mathcal{K}|=k}}{\arg\max}
    \sum_{f\in\mathcal{K}} \frac{\mathbf{c}_{t} \cdot \mathbf{c}_v^{f}}{\|\mathbf{c}_{t}\| \|\mathbf{c}_v^{f}\|}
\end{align} 
and the selected frames are those with the highest cosine similarity. Here, we directly select only the frames with the highest cosine similarity to a given text as a proxy for semantic similarity. Only the top-$k$ most semantically similar frames to a given text are pooled while lower similarity frames are completely ignored.

We observe that even by just applying top-$k$ pooling, there is already a significant improvement over baselines where the temporal aggregation function is text-agnostic. Detailed experiments can be found in Section \ref{sec:results}. 
\subsection{Our Model: \ModelName{}}
\label{sec:our-model}
\textbf{Towards Parametric Text-Conditioned Pooling.} However, there are still drawbacks with the top-$k$ method. Firstly, the tuning of the $k$ hyperparameter can be task and instance specific as we show in Section \ref{sec:results}. Secondly, deciding which frames to aggregate from can require more complex reasoning than simple cosine similarity. Lastly, completely suppressing frames with lower similarity may be too restrictive. As such, we propose a parametric approach to address these additional considerations while incorporating our insights from applying text-conditioned pooling.

\textbf{Cross-Modal Language-Video Attention.} Our idea is to design a learned frame aggregation function with parametric capacity for cross-modal reasoning about a text's most semantically similar frames in a video, which we call \ModelName{}. The core mechanism is our adaptation of a scaled dot product attention \cite{vaswani2017attention} between a text and the frames of a video. Conditioned on these frames, we generate a video embedding that learns to capture the most semantically similar video sub-regions as described in a given text. Since the frames with highest semantic similarity can differ depending on the text, our scaled dot product attention mechanism can learn to highlight relevant frames to a given text while suppressing frames not described in said text. Our model's capacity to selectively pick frames based on relevance to a given text is motivated by the same text-conditioning insights as outlined in the previously described top-$k$ approach. However, unlike the top-$k$ approach, our proposed model learns the optimal amount of information to extract for a text-video pair, thereby removing the need to manually specify a $k$ value. Furthermore, our cross-attention module handles both high and low relevancy frames rather than adopting a hard selection of relevant frames as in the top-$k$ approach.

To elaborate, in our cross-modal attention module, we first project a text embedding $\mathbf{c}_t \in \mathbb{R}^{D}$ into a single query $Q_t \in \mathbb{R}^{1 \times D_p}$ and a video's frame embeddings $C_v \in \mathbb{R}^{F \times D}$ into key $K_v \in \mathbb{R}^{F \times D_p}$ and value $V_v \in \mathbb{R}^{F\times D_p}$ matrices, where $D$ is the size of our model's latent dimension and $D_p$ is the size of the projection dimension. The projections are defined as:
\begin{equation}
    Q_t = \text{LN}(\mathbf{c}_t^T)W_Q
\end{equation}
\begin{equation}
    K_v = \text{LN}(C_v)W_K
\end{equation}
\begin{equation}
    V_v = \text{LN}(C_v)W_V 
\end{equation}
where LN is a Layer Normalization layer \cite{ba2016layer} and $W_Q$, $W_K$ and $W_V$ are projection matrices in $\mathbb{R}^{D \times D_p}$. In order to learn flexible conditioning between the given text and the frames, we then adapt scaled dot product attention from the query-projected text embedding to the key-projected frame embeddings. The dot product attention gives relevancy weights from a text to each frame which we leverage to aggregate the value-projected frame embeddings: 
\begin{equation}
    \text{Attention}(Q_t, K_v, V_v) = \text{softmax}\left(\frac{Q_t K_v^T}{\sqrt{D_p}}\right) V_v
\end{equation}

As such, the $Q_t$, $K_v$ and $V_v$ matrices can be interpreted akin to those in the original scaled dot product attention proposed in \cite{vaswani2017attention} except with cross-modal interactions. That is, the query-projected text embedding is used to seek from the key-projected frame embeddings to attend to frames with highest relevance. The value-projected embeddings represent the video's context from which we want to aggregate only certain sub-regions depending on the text. 

To embed a video into a joint space with a text, we project the aggregated video representation from the attention module back into $\mathbb{R}^D$ by applying a weight $W_O \in \mathbb{R}^{D_p \times D}$ to obtain:
\begin{equation}
   \mathbf{r}_{v \mid t} = \text{LN}(\text{Attention}(Q_t, K_v, V_v)W_O)
   \end{equation}
where the resulting output $\mathbf{r}_{t \mid v}$  is an aggregated video embedding conditioned on the text $t$. We can thereby learn this embedding such that a text can attend to its most semantically similar frames through parametric reasoning in the dot product attention. Our final text-conditioned pooling is defined as:
\begin{equation}
    \pi_{\text{\ModelName{}}}(C_v \mid t) = \text{LN}(\text{FC}(\mathbf{r}_{v \mid t})) + \mathbf{r}_{v \mid t})^T
\end{equation}
where FC is a fully connected network which together  with the residual connection provides additional capacity for more complex reasoning in our aggregation function.

Figure \ref{fig:arch} shows a diagram of our model. We show how \ModelName{} performs text-conditioned video aggregation over frames by allowing a text to learn to attend to its most semantically similar frames for pooling. In the top example, the input text $t_1$ is most relevant to the first few frames displayed of video $v_1$ of a man yelling at and punching a sink, whereas the final displayed frames of a man near a car do not capture what is described in the text and instead act as misleading visual distractors. We show how our model can reason about semantic similarity by assigning higher attention weights to the text's most relevant frames for aggregation. We emphasize that any text-agnostic pooling method such as mean-pooling would have aggregated the contents from this entire video. The resulting aggregation would thereby capture noisy distractors not described in the input text which could hamper the similarity score for retrieval. In the bottom example, we show a similar behaviour wherein \ModelName{} can attend to the most relevant frames of two guys jumping in an elevator as described in the text, whereas text-agnostic methods would capture non-relevant content from this video.   

\textbf{Loss.}
We train models using a dataset $\mathcal{D}$ consisting of $N$ text and video pairs $\{(t_i,v_i)\}_{i=1}^{N}$. In each pair, the text $t_i$ is a matching text description of the corresponding video $v_i$. We employ the cross entropy loss from \cite{zhai2018classification} by considering matching text-video pairs as positives and by considering all other pairwise text-video combinations in the batch as negatives. Specifically, we jointly minimize the symmetric text-to-video and video-to-text losses:
\begin{equation}
    \mathcal{L}_{t2v} = -\frac{1}{B} \sum_{i=1}^{B}{\text{log}\frac{e^{s(t_i,v_i) \cdot \lambda}}{\sum_{j=1}^{B}{e^{s(t_i,v_j) \cdot \lambda}}}} \\
    \label{eq:loss1}
\end{equation}
\begin{equation}
    \mathcal{L}_{v2t} = -\frac{1}{B} \sum_{i=1}^{B}{\text{log}\frac{e^{s(t_i,v_i) \cdot \lambda}}{\sum_{j=1}^{B}{e^{s(t_j,v_i) \cdot \lambda}}}}\\
    \label{eq:loss2}
\end{equation}
\begin{equation}
    \mathcal{L} = \mathcal{L}_{t2v} + \mathcal{L}_{v2t}
    \label{eq:loss}
\end{equation}
where $s(t_i,v_j)$ is the cosine similarity between the text $t_i$ and the video $v_j$, $B$ is the batch size and $\lambda$ is a learnable scaling parameter. By bootstrapping from a pre-trained CLIP model and through our cross-modal attention mechanism, training with this loss enables our model to learn to match a text with its most semantically similar sub-regions of the ground-truth video.

\section{Experiments}
We perform experiments on the commonly used benchmark text-video retrieval datasets of MSR-VTT \cite{xu2016msr}, MSVD \cite{chen2011collecting} and LSMDC \cite{rohrbach2017movie} and evaluate our performance following existing literature \cite{yu2018joint, miech2018learning, liu2019use,  gabeur2020multi, bain2021frozen} by reporting Recall@1 (R@1), Recall@5 (R@5), Recall@10 (R@10), Median Rank (MdR), and Mean Rank (MnR).

\subsection{Datasets}
\textbf{MSR-VTT} is comprised of 10,000 videos, each paired with about 20 human-labeled captions. We note that the multiple captions for each video in MSR-VTT often describe different video sub-regions, which supports our motivation for matching a given text with its most relevant frames in a video. The lengths of videos in this dataset range from 10 to 32 seconds, and we use two training splits which we call \textit{7k-Train} and \textit{9k-Train} to effectively compare with previous works. \textit{7k-Train} is a subset of roughly 7k videos as defined in \cite{miech2019howto100m}, while \textit{9k-Train} consists of approximately 9k videos following the split in \cite{gabeur2020multi}. Unless otherwise stated, we use the \textit{9k-Train} split for training. To evaluate our models, we use the \textit{1K-A} test set from \cite{yu2018joint} consisting of 1,000 selected caption-video pairs. 

\textbf{MSVD} contains about 120k captions that each describe one of 1,970 videos ranging in length from 1 to 62 seconds. Again, videos are paired with multiple captions and each may describe different sub-regions of the same video. In MSVD, the training, validation and test splits are comprised of 1,200, 100 and 670 videos respectively. Our final results are evaluated on the test split that has a varying number of captions per video. To that end, we follow recent methods for evaluation by treating all the provided caption-video pairs as separate instances for evaluation \cite{portillo2021straightforward, luo2021clip4clip}.

\textbf{LSMDC} is a movie clip dataset containing 118,081 videos each paired with a single caption description. The lengths of videos range from 2 to 30 seconds. 101,079 videos are used for training while 7,408 and 1,000 videos are used for validation and testing respectively. We report all results on the test set. 
 
\subsection{Implementation Details}
We use CLIP's ViT-B/32 image encoder as $\phi$ and CLIP's transformer base text encoder as $\psi$, and initialize all encoder parameters from CLIP's pre-trained weights. We set the query, key and value projection dimension size as $D_p=512$ to match CLIP's output dimension and initialize our logit scaling paramter $\lambda$ with that from a pre-trained CLIP model. We apply a linear layer with $D=512$ output units and dropout \cite{srivastava2014dropout} of 0.3 as our FC. Finally, we initialize all new projection weight matrices with identity and all new biases with zeros to bootstrap our entire model from the existing text-image semantic reasoning of a pre-trained CLIP. Our models are fine-tuned end-to-end on each dataset. To that end, we set our batch size to 32 for all experiments and set the learning rate for CLIP-initialized weights to 1e-6 and for all other parameters to 1e-5. We optimize our model for 5 epochs using the AdamW optimizer \cite{loshchilov2017decoupled} with weight decay set to 0.2 and decay the learning rate using a
cosine schedule \cite{loshchilov2016sgdr} following CLIP \cite{radford2021learning}. For all experiments, we uniformly sample 12 frames from every video and resize each frame to 224x224 following previous works \cite{liu2019use, bain2021frozen, luo2021clip4clip}.

\begin{table}[b]
\setlength{\tabcolsep}{2pt}
\centering
\footnotesize
\begin{tabular}{l c c c c c} 
\hline 
Methods & R@1 $\uparrow$ & R@5 $\uparrow$ & R@10 $\uparrow$ & MdR $\downarrow$ & MnR $\downarrow$ \\
\hline
CE \cite{liu2019use} & 20.9 & 48.8 & 62.4 & 6.0 & 28.2 \\
MMT \cite{gabeur2020multi} & 26.6 & 57.1 & 69.6 & 4.0 & 24.0 \\
Straight-CLIP \cite{portillo2021straightforward} & 31.2 & 53.7 & 64.2 & 4.0 & - \\
Support Set \cite{patrick2020support} & 30.1 & 58.5 & 69.3 & 3.0 & - \\
MDMMT \cite{dzabraev2021mdmmt} & 38.9 & 69.0 & 79.7 & \textbf{2.0} & 16.5 \\
Frozen \cite{bain2021frozen} & 31.0 & 59.5 & 70.5 & 3.0 & - \\
TeachText-CE+ \cite{croitoru2021teachtext} & 29.6 & 61.6 & 74.2 & 3.0 & - \\ 
CLIP4Clip-meanP \cite{luo2021clip4clip} & 43.1 & 70.4 & 80.8 & \textbf{2.0} & 16.2 \\
CLIP4Clip-seqTransf \cite{luo2021clip4clip} & 44.5 & 71.4 & 81.6 & \textbf{2.0} & 15.3 \\

\ModelName{} (ours) & \textbf{46.9} & \textbf{72.8} & \textbf{82.2} & \textbf{2.0} & \textbf{14.3} \\
\hline
\end{tabular}
\vspace{-0.2cm}
\caption{$t2v$ results on the MSR-VTT-9K dataset.}
\label{tab:msrvtt-9k-res}
\end{table}

\begin{table}[b]
\setlength{\tabcolsep}{2pt}
\centering
\footnotesize
\begin{tabular}{l c c c c c} 
\hline 
Methods & R@1 $\uparrow$ & R@5 $\uparrow$ & R@10 $\uparrow$ & MdR $\downarrow$ & MnR $\downarrow$ \\
\hline
HowTo100M \cite{miech2019howto100m} & 14.9 & 40.2 & 52.8 & 9.0 & - \\
ActBERT \cite{zhu2020actbert} & 8.6 & 23.4 & 33.1 & 36.0 & - \\
NoiseE \cite{amrani2020noise} & 17.4 & 41.6 & 53.6 & 8.0 & - \\
ClipBERT \cite{lei2021less} & 22.0 & 46.8 & 59.9 & 6.0 & - \\
CLIP4Clip-meanP \cite{luo2021clip4clip} & 42.1 & 71.9 & 81.4 & \textbf{2.0} & 15.7 \\
CLIP4Clip-seqTransf \cite{luo2021clip4clip} & 42.0 & 68.6 & 78.7 & \textbf{2.0} & 16.2 \\

\ModelName{} (ours) & \textbf{43.9} & \textbf{72.5} & \textbf{82.3} & \textbf{2.0} & \textbf{14.6} \\

\hline
\end{tabular}
\vspace{-0.2cm}
\caption{$t2v$ results on the MSR-VTT-7K dataset.}
\label{tab:msrvtt-7k-res}
\end{table}

\subsection{Results}
\label{sec:results}
To evaluate our method, we compare its performance with recent works from the literature. We tabulate the $t2v$ retrieval performance of our model trained on the MSR-VTT \textit{9k-Train} and \textit{7k-Train} splits in Table \ref{tab:msrvtt-9k-res} and Table \ref{tab:msrvtt-7k-res} respectively. Tables \ref{tab:msvd-res} and \ref{tab:lsmdc-res} similarly compare the performance of \ModelName{} on the MSVD and LSMDC datasets respectively. We note that on all datasets and across all metrics, our text-conditioned \ModelName{} model outperforms all other works that use text-agnostic pooling \cite{bain2021frozen, portillo2021straightforward, luo2021clip4clip} including those using video experts in multiple video modalities \cite{gabeur2020multi, miech2020end, alayrac2020self}. Most notably, our model outperforms the hitherto state-of-the-art methods CLIP4Clip-meanP and CLIP4Clip-seqTransf \cite{luo2021clip4clip} which are the most directly comparable to \ModelName{} since they also use CLIP as a backbone. Therefore, we can directly attribute the performance gains of our model to the fact that we use text-conditioned pooling compared to the text-agnostic pooling schemes of CLIP4Clip-meanP and CLIP4Clip-seqTransf.

More precisely, on the MSR-VTT dataset, we observe a relative improvement of 5\% in Recall@1 compared to CLIP4Clip-seqTransf. For the MSVD dataset, we outperform CLIP4Clip-meanP by over 2\% in relative improvement in Recall@1. In the case of the LSMDC dataset, the retrieval problem is more challenging since the movie scene text descriptions are much more ambiguous, which can be observed by the overall lower retrieval scores of all previous methods. Yet, our method notably outperforms CLIP4Clip-seqTransf by 12\% in relative improvement in Recall@1. Our results thereby highlight the importance of our model's text-conditioned aggregation that can learn to match a text with its most relevant frames while suppressing distracting visual cues from other video sub-regions.

\begin{table}[]
\footnotesize
\setlength{\tabcolsep}{2pt}
\centering
\begin{tabular}{l c c c c c} 
\hline 
Methods & R@1 $\uparrow$ & R@5 $\uparrow$ & R@10 $\uparrow$ & MdR $\downarrow$ & MnR $\downarrow$ \\
\hline
CE \cite{liu2019use} & 19.8 & 49.0 & 63.8 & 6.0 & 23.1 \\
Support Set \cite{patrick2020support} & 28.4 & 60.0 & 72.9 & 4.0 & - \\
NoiseE \cite{amrani2020noise} & 20.3 & 49.0 & 63.3 & 6.0 & - \\
Straight-CLIP \cite{portillo2021straightforward} & 37.0 & 64.1 & 73.8 & 3.0 & - \\
Frozen \cite{bain2021frozen} & 33.7 & 64.7 & 76.3 & 3.0 & - \\
TeachText-CE+ \cite{croitoru2021teachtext} & 25.4 & 56.9 & 71.3 & 4.0 & - \\ 
CLIP4Clip-meanP \cite{luo2021clip4clip} & 46.2 & 76.1 & 84.6 & \textbf{2.0} & 10.0 \\
CLIP4Clip-seqTransf \cite{luo2021clip4clip} & 45.2 & 75.5 & 84.3 & \textbf{2.0} & 10.3 \\
\ModelName{} (ours) & \textbf{47.2} & \textbf{77.4} & \textbf{86.0} & \textbf{2.0} & \textbf{9.3} \\
\hline
\end{tabular}
\vspace{-0.2cm}
\caption{$t2v$ results on the MSVD dataset.}
\label{tab:msvd-res}
\end{table}

\begin{table}[]
\footnotesize
\setlength{\tabcolsep}{2pt}
\centering
\begin{tabular}{l c c c c c} 
\hline 
Methods & R@1 $\uparrow$ & R@5 $\uparrow$ & R@10 $\uparrow$ & MdR $\downarrow$ & MnR $\downarrow$ \\
\hline
CE \cite{liu2019use} & 11.2 & 26.9 & 34.8 & 25.3 & - \\
MMT \cite{gabeur2020multi} & 12.9 & 29.9 & 40.1 & 19.3 & 75.0 \\
NoiseE \cite{amrani2020noise} & 6.4 & 19.8 & 28.4 & 39.0 & - \\
Straight-CLIP \cite{portillo2021straightforward} & 11.3 & 22.7 & 29.2 & 56.5 & - \\
MDMMT \cite{dzabraev2021mdmmt} & 18.8 & 38.5 & 47.9 & 12.3 & 58.0 \\
Frozen \cite{bain2021frozen} & 15.0 & 30.8 & 39.8 & 20.0 & - \\
TeachText-CE+ \cite{croitoru2021teachtext} & 17.2 & 36.5 & 46.3 & 13.7 & - \\
CLIP4Clip-meanP \cite{luo2021clip4clip} & 20.7 & 38.9 & 47.2 & 13.0 & 65.3 \\
CLIP4Clip-seqTransf \cite{luo2021clip4clip} & 22.6 & 41.0 & 49.1 & 11.0 & 61.0 \\
\ModelName{} (ours) & \textbf{25.2} & \textbf{43.7} & \textbf{53.5} & \textbf{8.0} & \textbf{53.2} \\
\hline
\end{tabular}
\vspace{-0.2cm}
\caption{$t2v$ results on the LSMDC dataset.}
\label{tab:lsmdc-res}
\end{table}

\textbf{Top-$\boldsymbol{k}$ Experiments.}
To better understand the merits and intuition for our \ModelName{} model, we first revisit our top-$k$ temporal aggregation function defined in equation \eqref{eq:topk} that we introduce as a proof of concept for our proposed idea of text-conditioned video pooling. To validate this idea, we compare top-$k$ pooling with a mean-pooling baseline as in \cite{portillo2021straightforward, luo2021clip4clip} across two settings: first we apply a pre-trained CLIP model in a zero-shot manner similar to \cite{portillo2021straightforward} to compare mean-pooling and top-$k$ aggregation, and second we fine-tune a pre-trained CLIP model on the MSR-VTT dataset and then measure retrieval performance for mean-pooling and top-$k$ pooling. In both settings, we set $k$=3 which empirically yields the best overall performance. We compare the $t2v$ results in Table \ref{fig:poc_tab} and observe that even by using cosine similarity in top-$k$ pooling as a proxy for semantic similarity between a text and frames, we can outperform mean-pooling across all listed metrics by up to 6\% of relative improvement in Recall@1 through our text-conditioned pooling scheme. 

Yet, the top-$k$ aggregation function still presents some drawbacks as mentioned in Section \ref{sec:no-text-pool}, most notably relating to the tuning of the $k$ hyperparameter. To analyze this shortcoming, we run an experiment wherein for a zero-shot pre-trained CLIP, we find the optimal $k$ of each individual text-video pair in the MSR-VTT test set and report the results in a histogram in Figure \ref{fig:poc_k}. Here, we define optimal as the the $k$ value that yields the highest similarity score between a ground-truth text-video pair as defined in equation \eqref{eq:cosine-sim-new}. We observe that the optimal choice of $k$ varies widely between text-video pairs, which makes $k$ difficult to select in general. Our proposed \ModelName{} model therefore addresses the drawbacks of top-$k$ pooling while being motivated by our derived insights of text-conditioned pooling.

\setlength\fboxrule{0pt}
\begin{figure}[tb]\centering
	\setlength\tabcolsep{0pt} 
	\begin{tabular}{cc}
        \resizebox{0.65\columnwidth}{!}{
			\centering\fbox{
			    \setlength{\tabcolsep}{2pt}
                \centering
			    \begin{tabular}{l c c c c c} 
                \hline 
                Aggr. & R@1 $\uparrow$ & R@5 $\uparrow$ & R@10 $\uparrow$ & MdR $\downarrow$ & MnR $\downarrow$ \\
                \hline
                \multicolumn{6}{c}{Zero-Shot CLIP} \\
                \hline
                Mean & 31.5 & 52.8 & 63.6 & 5.0 & 42.9 \\
                Top-$k$ & \textbf{33.6} & \textbf{54.0} & \textbf{64.3} & \textbf{4.0} & \textbf{42.5} \\
                \hline
                \multicolumn{6}{c}{Fine-Tuned CLIP} \\
                \hline
                Mean & 42.1 & 69.8 & 80.7 & \textbf{2.0} & 15.7 \\
                Top-$k$ & \textbf{44.6} & \textbf{70.9} & \textbf{82.4} & \textbf{2.0} & \textbf{14.9} \\
                \hline
                \end{tabular}
			}
		}
		&
		\resizebox{0.35\columnwidth}{!}{
			\centering\fbox{
			\hspace{-0.5cm}
			\begin{tikzpicture}[baseline=1cm]
                \LARGE
                \begin{axis}[ybar interval, height=5cm, ymax=450, ymin=0, grid=major, xmin=0.75, xmax=7.25, xticklabels={1, 3, 5, 7, 9, 11}, xlabel={$k$}, ylabel={\# of Pairs}, ylabel near ticks, xlabel near ticks, major tick length=0]
                \addplot coordinates { (1, 166) (2, 401) (3, 204) (4, 98) (5, 55) (6, 76) (7,0) };
                \end{axis}
            \end{tikzpicture}
			}
		} \\[-0.5cm] \subfloat[\label{fig:poc_tab}]{\quad}   &
        \subfloat[\label{fig:poc_k}]{\quad}\\[-0.3cm]
	\end{tabular}
	\caption{Top-$k$ analysis on MSR-VTT.
	\protect\subref{fig:poc_tab} $t2v$ retrieval performance comparing mean-pooling with top-$k$ text-conditioned pooling.
        \protect\subref{fig:poc_k}
        Histogram showing the $k$ value where each ground truth text-video pair in the MSR-VTT test set achieves the highest cosine similarity when using top-$k$ pooling.
        }\label{fig:poc}
	
\end{figure}
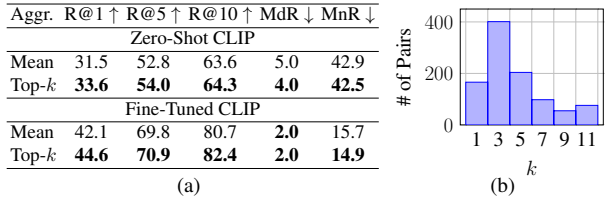

\begin{figure}[t]
\centering
\begin{tikzpicture}
\small
\begin{axis}[width=\columnwidth, height=3.5cm, ymax=50, ymin=0, grid=major, xmin=0.75, xmax=5.25, xtick=data, xticklabels={0, 1, 2, 3, 4}, xlabel={Number of Transitions}, ylabel={Median Rank}, ylabel near ticks, xlabel near ticks, legend style={font=\small}, legend pos=north west,legend cell align={left}]
\addplot coordinates {(1, 2) (2, 3) (3, 4) (4, 6) (5, 9)};\addlegendentry{\ModelName{}}
\addplot coordinates {(1, 2) (2, 8) (3, 18) (4, 26) (5, 46)};\addlegendentry{Mean-Pooling}
\end{axis}
\end{tikzpicture}
\caption{Robustness to content diversity. We show the $t2v$ Median Rank results on MSR-VTT for different amounts of content diversity measured by the number of scene transitions. Our \ModelName{} approach remains robust whereas mean-pooling significantly deteriorates as we increase the content diversity.}
\label{fig:augmentation_exp}
\end{figure}
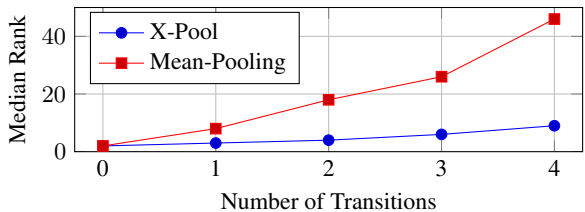

\textbf{Robustness to Content Diversity in Videos.} 
We now analyze the robustness of our model to content diversity as we described in Section \ref{sec:no-text-pool}. As explained, many videos inherently exhibit diverse visual content such as scene transitions or changes in object appearance for example. While current datasets such as MSR-VTT, LSMDC and MSVD already display these traits to an extent, they
are curated by choosing only small video clip segments extracted from larger videos. 
Therefore, in order to more effectively test the robustness of text-video retrieval methods to content diversity, one way is to 
introduce additional diversity in visual content with more scene transitions. That is, we augment a video's visual content by randomly injecting another video from the dataset to simulate an abrupt scene transition.  By performing retrieval on such augmented videos and their original text captions, we can better evaluate a retrieval model's ability to handle diverse videos in the wild.

To that end, we construct augmented versions
of the MSR-VTT test set by adding scene transitions from each video to other videos in the test set.
The number of transitions is defined as the number of random videos
that are added to the original video at a random location. We compare the $t2v$ retrieval performance of our \ModelName{} model to the baseline of mean-pooling, and plot the results in Figure \ref{fig:augmentation_exp}. Here, we measure performance using the metric of Median Rank. We can clearly observe
that as the number of video transitions increases and we add video content diversity, there is a sharp performance decline in mean-pooling as the Median Rank increases
from 2 to 46, whereas our \ModelName{} model is significantly more robust to content diversity as Median rank only increases from 2 to 9. The performance gap is because any text-agnostic pooling method like mean-pooling aggregates content from all scenes of a video regardless of their relevance to an input text. Therefore, the more diverse a video is in terms of scene transitions, the more possibly noisy distractors are being aggregated. Conversely, \ModelName{} can extract only the most relevant visual cues as described in a text through text-conditioned pooling.

\textbf{Qualitative Results.}
In Figure \ref{fig:augmentation_exp}, we show qualitative examples of our \ModelName{} model. For each example, we show four sampled frames from a video along with a bar plot representing the associated attention weights of \ModelName{} from the given text to each frame. In the top example, we can see that our model outputs a higher attention weight for the middle frames when the input text describes a brain animation and lower attention weights everywhere else. On the other hand, when the input text instead describes a fictional character looking at a machine, the attention weight correspondingly activates for the last frame where the text is most relevant. The second example in the middle shows a singing competition. Here, the text of ``a judge hearing the voice of competitors'' describes an event that requires reasoning over all of the frames. Indeed, we observe that \ModelName{} attends to the entire video, indicating the flexibility of our approach. Finally, in the bottom example, we observe that our model correspondingly activates on the most relevant frames to each text despite the more subtle nuances in the language and video semantics of this lion example.

\begin{figure}[t]\centering \includegraphics[width=0.5\textwidth]{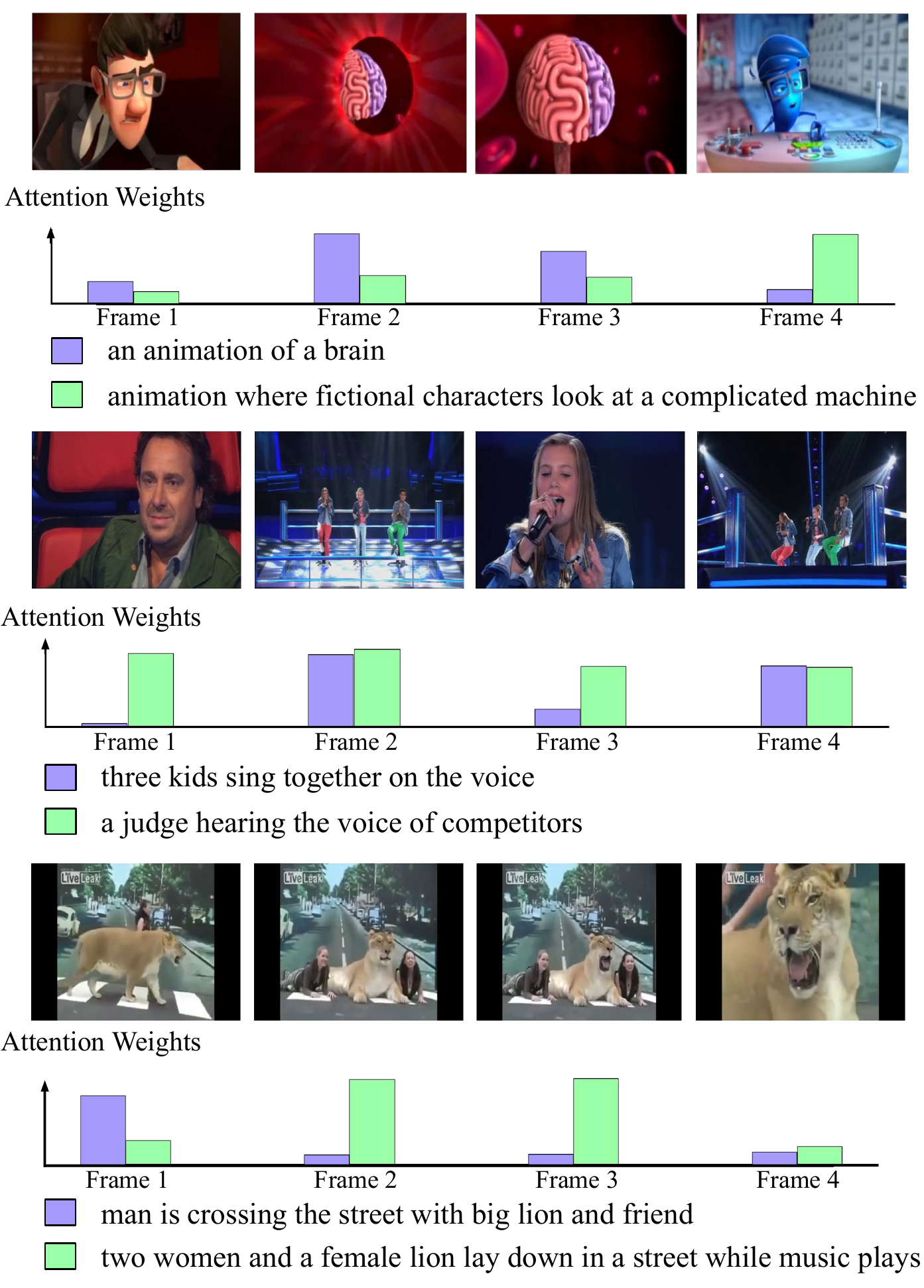}\\
  \caption{Qualitative results of \ModelName{} from the MSR-VTT dataset. For each displayed frame above, the bar plot shows its attention weights in our model given a particular text.
  }
\label{fig:example3}
\end{figure}

\section{Conclusion}
In this work, we highlight the drawbacks of text-agnostic video pooling and present an alternative framework for text-conditioned pooling for text-video retrieval.  We then extend our idea and derived insights to design a parametric model for cross-modal attention between a text and video frames called \ModelName{}. We show how \ModelName{} can learn to attend to the most relevant frames to a given a text, which also makes our model substantially more robust to video content diversity such as in the form of scene transitions, a property that is common in videos in the wild. As part of future work, we plan on applying text-conditioned video pooling to other cross-modal tasks like video question answering. 

{\small
\bibliographystyle{ieee_fullname}
\bibliography{egbib}
}

\newpage
\appendix
\appendixpage
\counterwithin{figure}{section}
\counterwithin{table}{section}
\renewcommand\thefigure{\thesection\arabic{figure}}
\renewcommand\thetable{\thesection\arabic{table}}
\section{Video-to-Text Retrieval Results}

\begin{table}[h]
\setlength{\tabcolsep}{2pt}
\centering
\footnotesize
\begin{tabular}{l c c c c c} 
\hline 
Methods & R@1 $\uparrow$ & R@5 $\uparrow$ & R@10 $\uparrow$ & MdR $\downarrow$ & MnR $\downarrow$ \\
\hline
CE \cite{liu2019use} & 20.6 & 50.3 & 64.0 & 5.3 & 25.1 \\
MMT \cite{gabeur2020multi} & 27.0 & 57.5 & 69.7 & 3.7 & 21.3 \\
Straight-CLIP \cite{portillo2021straightforward} & 27.2 & 51.7 & 62.6 & 5.0 & - \\
Support Set \cite{patrick2020support} & 28.5 & 58.6 & 71.6 & 3.0 & - \\
TeachText-CE+ \cite{croitoru2021teachtext} & 32.1 & 62.7 & 75.0 & 3.0 & - \\ 
CLIP4Clip-meanP \cite{luo2021clip4clip} & 43.1 & 70.5 & 81.2 & \textbf{2.0} & 12.4 \\
CLIP4Clip-seqTransf \cite{luo2021clip4clip} & 42.7 & 70.9 & 80.6 & \textbf{2.0} & 11.6 \\
\ModelName{} (ours) & \textbf{44.4} & \textbf{73.3} & \textbf{84.0} & \textbf{2.0} & \textbf{9.0} \\
\hline
\end{tabular}
\vspace{-0.2cm}
\caption{$v2t$ results on the MSR-VTT-9K dataset.}
\label{tab:msrvtt-9k-res-v2t}
\end{table}

\begin{table}[h]
\footnotesize
\setlength{\tabcolsep}{2pt}
\centering
\begin{tabular}{l c c c c c} 
\hline 
Methods & R@1 $\uparrow$ & R@5 $\uparrow$ & R@10 $\uparrow$ & MdR $\downarrow$ & MnR $\downarrow$ \\
\hline
Straight-CLIP \cite{portillo2021straightforward} & 59.9 & 85.2 & 90.7 & \textbf{1.0} & - \\
TeachText-CE+ \cite{croitoru2021teachtext} & 27.1 & 55.3 & 67.1 & 4.0 & - \\ 
CLIP4Clip-meanP \cite{luo2021clip4clip} & 56.6 & 79.7 & 84.3 & \textbf{1.0} & 7.6 \\
CLIP4Clip-seqTransf \cite{luo2021clip4clip} & 62.0 & 87.3 & 92.6 & \textbf{1.0} & 4.3 \\
\ModelName{} (ours) & \textbf{66.4} & \textbf{90.0} & \textbf{94.2} & \textbf{1.0} & \textbf{3.3} \\
\hline
\end{tabular}
\vspace{-0.2cm}
\caption{$v2t$ results on the MSVD dataset.}
\label{tab:msvd-res-v2t}
\end{table}

\begin{table}[h]
\footnotesize
\setlength{\tabcolsep}{2pt}
\centering
\begin{tabular}{l c c c c c} 
\hline 
Methods & R@1 $\uparrow$ & R@5 $\uparrow$ & R@10 $\uparrow$ & MdR $\downarrow$ & MnR $\downarrow$ \\
\hline
JSFusion \cite{yu2018joint} & 12.3 & 28.6 & 38.9 & 20.0 & - \\
Straight-CLIP \cite{portillo2021straightforward} & 6.8 & 16.4 & 22.1 & 73.0 & - \\
TeachText-CE+ \cite{croitoru2021teachtext} & 17.5 & 36.0 & 45.0 & 14.3 & - \\
CLIP4Clip-meanP \cite{luo2021clip4clip} & 20.6 & 39.4 & 47.5 & 13.0 & 56.7 \\
CLIP4Clip-seqTransf \cite{luo2021clip4clip} & 20.8 & 39.0 & 48.6 & 12.0 & 54.2 \\
\ModelName{} (ours) & \textbf{22.7} & \textbf{42.6} & \textbf{51.2} & \textbf{10.0} & \textbf{47.4} \\
\hline
\end{tabular}
\vspace{-0.2cm}
\caption{$v2t$ results on the LSMDC dataset.}
\label{tab:lsmdc-res-v2t}
\end{table}

\section{Number of Frames Experiment}
\begin{figure}[t]
\centering
\begin{tikzpicture}
\begin{axis}[width=\columnwidth, height=3.5cm, ymax=49, ymin=38, grid=major, xmin=0.75, xmax=4.25, xtick=data, xticklabels={6, 12, 24, All}, xlabel={Number of Frames}, ytick={40,44,48}, ylabel={Recall@1}, ylabel near ticks, xlabel near ticks, name=border]
\addplot coordinates {(1, 43.8) (2, 47.0) (3, 47.1) (4, 47.7)};\addlegendentry{\ModelName{}}\label{legend.A}
\addplot coordinates {(1, 39.4) (2, 42.1) (3, 41.4) (4, 41.9)};\addlegendentry{Mean-Pooling}\label{legend.B}
\legend{}; 
\end{axis}
\path let \p1=(border.east), \p2=(border.west), \n1={veclen(\x1-\x2,0)} in node[draw,inner xsep=0pt,above=0.1cm] at (border.north)   {\makebox[\n1]{\ref{legend.A} \ModelName{} \hfil\ref{legend.B} Mean-Pooling}};
\end{tikzpicture}
\vskip -0.3cm
\caption{$t2v$ Recall@1 results on the MSR-VTT-9K dataset when varying the number of frames. ``All'' indicates inference with all frames.}
\label{fig:frames_exp}
\vskip -0.4cm
\end{figure}
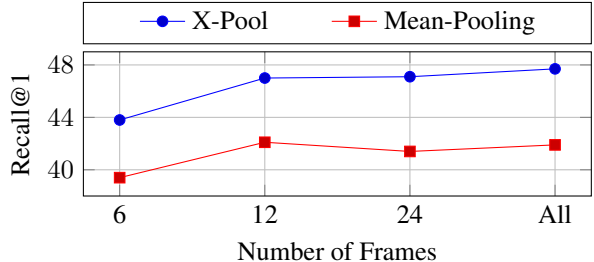

Our experiments use 12 sampled frames by default following recent text-video retrieval literature \cite{luo2021clip4clip}, and we run additional experiments on the MSR-VTT-9K dataset by varying the number of sampled frames for both training and inference as shown in Figure \ref{fig:frames_exp}. We observe worse performance for 6 frames likely due to important information being missing at this scale. As we increase the number of frames\footnote{"All" indicates inference with all frames at inference time after training on 12 sampled frames.}, we observe performance saturation which is consistent with findings in \cite{luo2021clip4clip}. However, we note that the optimal number of sampled frames remains a dataset specific hyperparameter.

\section{Online Inference in a Large-Scale Production System}
Since our model computes an aggregated video embedding conditioned on a given text, the embeddings from a video index set in $t2v$ cannot be entirely pre-computed because query texts are not a priori known during online inference. Instead, we can only pre-compute the frame embeddings of each index video, so fast nearest neighbour retrieval techniques \cite{liu2004investigation, johnson2019billion} cannot be readily applied. To address this in a production system with large-scale index sets, one commonly used approach is to use a high recall method to obtain a set of top retrieval candidates using using a nearest-neighbour search, and then use another method yielding high precision to re-rank the candidates \cite{covington2016deep, ma2019cross}. 

In our case, we can first mean-pool the pre-computed frame embeddings coming from \ModelName{} and then very efficiently obtain a set of $\mathcal{P}$ most similar candidates from the index set given a retrieval query. We can then run \ModelName{}’s text-conditioned attention mechanism only on said candidates and then re-rank them for retrieval. That way, given $\mathcal{T}$ text queries and $\mathcal{V}$ index videos in $t2v$, instead of an $\mathcal{O}(\mathcal{T}\mathcal{V})$ complexity, we can achieve an $\mathcal{O}(\mathcal{T}\mathcal{P} + \mathcal{V})$ complexity where $\mathcal{P} << \mathcal{V}$ while maintaining good performance. In fact, we evaluated the performance of our model on the MSR-VTT dataset using the top-100 candidates from mean-pooling (i.e. $\mathcal{P}=100$) and obtained the same performance in Recall@1, Recall@5 and Recall@10 as listed in our main results. 

\end{document}